\pdfoutput=1

\documentclass[11pt]{article}

\usepackage{coling}

\usepackage{pdfpages}
\usepackage{times}
\usepackage{latexsym}

\usepackage[T1]{fontenc}

\usepackage[utf8]{inputenc}
\usepackage{cleveref}
\usepackage{graphicx} 

\usepackage{soulutf8}
\usepackage{array}

\usepackage{xcolor}

\usepackage{color-edits}
\addauthor{z}{magenta}

\usepackage{microtype}

\usepackage{inconsolata}

\usepackage{todonotes}

\usepackage{multicol} 
\usepackage{cuted}
\usepackage{enumitem}

\makeatletter
\newcommand*\iftodonotes{\if@todonotes@disabled\expandafter\@secondoftwo\else\expandafter\@firstoftwo\fi}  

\usepackage{multirow}
\usepackage{url}

\usepackage{arydshln}
\usepackage[T1]{fontenc}

\title{From Multiple-Choice to Extractive QA:\\ A Case Study for English and Arabic}

\author{Teresa Lynn,\textsuperscript{1} Malik H. Altakrori,\textsuperscript{2} \textbf{Samar M. Magdy}\textsuperscript{1} \textbf{Rocktim Jyoti Das},\textsuperscript{1} \\
\textbf{Chenyang Lyu},\textsuperscript{1} \textbf{Mohamed Nasr},\textsuperscript{2} \textbf{Younes Samih},\textsuperscript{2} 
\textbf{Kirill Chirkunov},\textsuperscript{1} 
\textbf{Alham Fikri Aji},\textsuperscript{1} \\ \textbf{Preslav Nakov},\textsuperscript{1} \textbf{Shantanu Godbole},\textsuperscript{2} \textbf{Salim Roukos},\textsuperscript{2} \textbf{Radu Florian},\textsuperscript{2} \textbf{Nizar Habash}\textsuperscript{1,3} \\
 $^1$Mohamed bin Zayed University of Artificial Intelligence\\
 $^2$IBM Research AI \quad
  $^3$New York University Abu Dhabi\\
  \texttt{teresa.lynn@mbzuai.ac.ae, malik.altakrori@ibm.com, nizar.habash@nyu.edu}
  }
  
\begin{document}
\maketitle
\begin{abstract}

The rapid evolution of Natural Language Processing (NLP) has favoured major languages such as English, leaving a significant gap for many others due to limited resources. 
This is especially evident in the context of data annotation, a task whose importance cannot be underestimated, but which is time-consuming and costly.
%
Thus, any dataset for resource-poor languages is precious, in particular when it is task-specific.
Here, we explore the feasibility of repurposing an existing multilingual dataset for a new NLP task: 
we repurpose a subset of the {\sc Belebele} dataset ~\cite{bandarkar2023belebele}, which was designed for \textit{multiple-choice} question answering (MCQA), to enable the more practical task of \textit{extractive} QA (EQA) in the style of machine reading comprehension. 
We present annotation guidelines and a parallel EQA dataset for English and Modern Standard Arabic (MSA).  
We also present QA evaluation results for several monolingual and cross-lingual QA pairs including English, MSA, and five Arabic dialects.
We aim to help others adapt our approach for the remaining 120  {\sc Belebele} language variants, many of which are deemed under-resourced. We also provide a thorough analysis and share insights to deepen understanding of the challenges and opportunities in NLP task reformulation.
\end{abstract}

\section{Introduction}
Recent years have brought about very fast developments in Natural Language Processing (NLP). However, this progress has not been equal for all languages, and most research has focused on English and a handful of other languages, with the vast majority of other languages being overlooked~\cite{joshi-etal-2020-state, Rehm2023-ele}. This is mainly due to the lack of data resources, which hampers the development of NLP tools for these languages. While many resources have been developed for some languages, they are often for very specific tasks, and for very specific formulations of these tasks. 

Starting from a motivation of creating cross-lingual Question Answering (QA) datasets for an under-resourced pair, Dialectal Arabic (DA) -- Modern Standard Arabic (MSA), we explore the possibility of re-purposing the multilingual {\sc Belebele} dataset~\cite{bandarkar2023belebele}, which was created for the task of multiple-choice question answering, to be suitable for extractive question answering in the style of machine reading comprehension.

In the field of Question Answering, there are a number of different ways to answer a question ~\cite{wang2022modern_mrc_qa_survey}, e.g., (a)~\textbf{multiple-choice QA} (\textbf{MCQA}) \textit{with answer choices, and context provided}, (b)~\textbf{extractive QA} (\textbf{EQA}) \textit{with context provided}, (c)~\textbf{abstractive QA} \textit{with context provided}, and (d)~\textbf{open-ended QA} \textit{with no context provided}. An interesting research question is whether we can traverse between them, i.e., if we have a resource for one QA task, can we repurpose it easily for a different reformulation of the task? Here we study this research question, in the context of transferring from MCQA to EQA, as we deem EQA more useful in real-world applications where the correct answer is often not known. Our contributions are:

\begin{itemize}
\vspace{-1pt}
\setlength{\itemsep}{-1pt}
\setlength{\parsep}{-1pt}
\setlength{\topskip}{-1pt}
    \item We explore the possibility for repurposing a MCQA dataset as an EQA dataset.
    \item We create and release a new parallel EQA dataset for MSA and English (EN).
    \item We provide a dialectal Arabic QA benchmark by evaluating our EQA system's performance on monolingual and cross-lingual QA pairs of 
    \{EN, MSA\}--\{EN, MSA, five DAs\}.
    \item We provide guidelines and a strong foundation for creating EQA datasets for the remaining 120 {\sc Belebele} languages.
    \item We perform careful analysis and discuss the lessons learned as a guide for future research in repurposing NLP datasets.

\end{itemize}





\section{Related Work}
\label{sec:related}


\subsection{Multilingual QA Datasets}
\label{sec:MultilingDatasets}
Numerous multilingual QA datasets have already been developed e.g., XQUAD~\cite{artetxe-etal-2020-cross}, TyDi QA~\cite{clark-etal-2020-tydi}, and MLQA~\cite{lewis-etal-2020-mlqa}, which focus on extractive-based QA. The recent ArabicaQA dataset~\cite{abdallah2024arabicaqa} represents a significant leap forward, being the first large-scale dataset specifically designed for Arabic QA: it has over 89k questions, derived from a diverse set of Standard Arabic documents, providing a robust platform for advancing Arabic NLP, particularly in the context of large language models.
Additionally, alternative formats of multilingual QA have been explored, including open-ended QA with Mintaka~\cite{sen-etal-2022-mintaka} and MCQA with {\sc Belebele}~\cite{bandarkar2023belebele}. With the exception of {\sc Belebele}, these datasets encompass a relatively narrow range of languages and do not include Arabic dialects. 
To address this gap, we propose leveraging the {\sc Belebele} dataset, to enable the inclusion of its dialectal content into a new parallel DA -- MSA/EN dataset.

\subsection{Converting MCQA Datasets}
\label{sec:ConvertingMCQA}
Some attempts have been made to convert existing datasets into a different format. Specifically for QA, there has been conversion of MCQA datasets \emph{into} natural language inference (NLI)~\cite{demszky2018transforming, khot2018scitail}: the premise is taken from the context paragraph, the correct choice is used as an entailment label, and the other choices are used as neutral or contradictory. Both approaches require effort to rephrase the question into a statement, e.g., using rule-based methods~\cite{demszky2018transforming} or human annotation~\cite{khot2018scitail}. \citet{hadifar2022eduqg} introduced a new dataset specifically designed for the educational setting, focusing on converting between multiple-choice and open-ended question formats.
Automatically generating MCQA has also been explored~\cite{mitkov-ha-2003-computer,kurdi2020systematic,ai-etal-2015-semi,karamanis-etal-2006-generating}, where the challenge is to generate distractor choices. 

To our knowledge, there is no previous work on converting MCQA to EQA datasets. In this paper, we investigate the feasibility of such a task with an eye towards future automation of the process.

    

\begin{table*}[h!]
    \centering
    \includegraphics[width=0.97\textwidth]{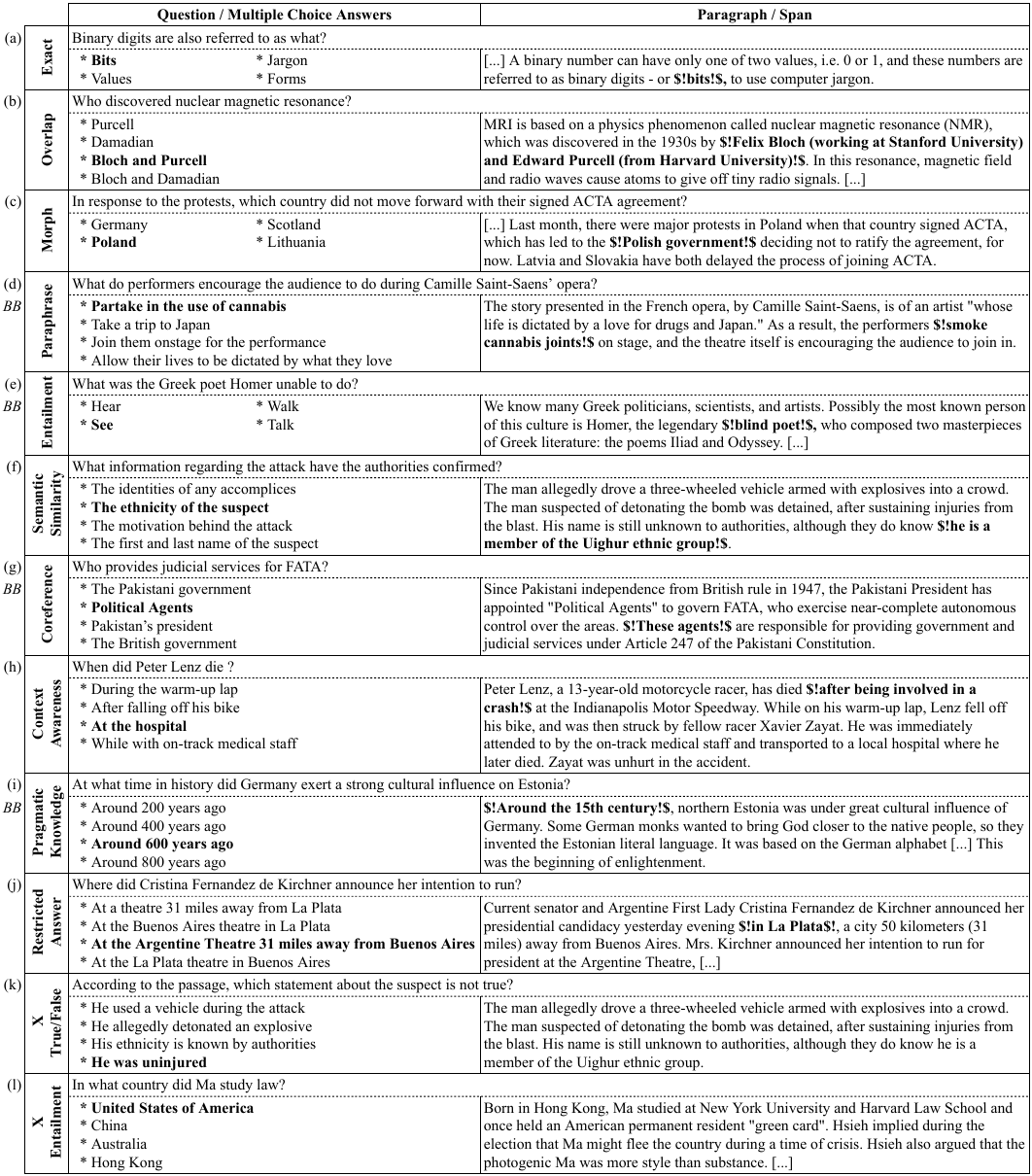}
    \caption{English examples demonstrating some interesting types of MCQs in the {\sc Belebele} dataset, and the respective issues they pose for extractive QA in this work. The Arabic version is shown in Appendix~\ref{app:ar-examples}. \vspace{10pt}}
    \label{tab:belebele_examples}
\end{table*}

\subsection{A Note on Arabic and its Dialects}
The Arabic language is a family of variants, among which Modern Standard Arabic (MSA) is the Arab world's shared language of culture, media, and education. However, MSA is not the native language of
any speaker of Arabic, whose day-to-day language is typically a local variety, i.e., Dialectal Arabic (DA) that can be quite different from MSA and other dialects \cite{salameh2018fine}.
MSA and DA coexist in what is called a diglossic relationship where different variants are used in different contexts \cite{ferguson1959diglossia}.
In the context of Arabic QA, there are many efforts  that focus primarily on MSA \cite{shaheen2014arabic,mozannar-etal-2019-neural,biltawi2021arabic,alwaneen2022arabic}.
A recent exception is the work of \newcite{faisal2021sd}, which works on spoken dialectal QA and includes Arabic among other languages.

The common wisdom in Arabic NLP is that systems should be robust to handling DA forms, but should generate output primarily in MSA. This is reasonable for QA in particular, given the relatively larger trusted content in MSA compared to DA which dominates social media.
As such, in this paper, we focus on modeling QA that accepts questions in a range of dialects and provides answers in standard languages (MSA and EN).
We selected the following five representative dialects from across the Arab world to work with:
Egyptian (arz),
Iraqi (acm),
Moroccan (ary),
Gulf (Najdi; ars), and 
Levantine (North; apc).

\section{MCQA-to-EQA Conversion Challenges}
\label{sec:challenges}

\subsection{Belebele Dataset~\label{sec:BBDataset}}

{\sc Belebele}~\cite{bandarkar2023belebele} is an open-source \textit{multiple-choice} machine reading comprehension (MRC) dataset that covers 122 languages and language variants, 
including English, MSA, and five Arabic dialects. Each question is based on a short passage from the Flores-200 dataset~\cite{nllb2022} and has four multiple-choice answers, with the correct answer identified. This benchmark dataset enabled the evaluation of natural language understanding (NLU) across high/medium/low-resource languages and language variants. 

With respect to the style of the dataset content, the creators of the {\sc Belebele} dataset indicated that they selected 
multiple-choice questions (MCQs) because \textit{it
would lead to the fairest evaluation across languages} and \textit{MCQs enable the ability
to scale to many languages when translating from English}.
They worked with a Language Service Provider 
to create the questions and the multi-choice answers in an iterative process, by developing guidelines that included rules such as \textit{no use of double negatives}, 
only use answers that \textit{are decently plausible to require the test-taker to fully read and understand the passage}~\cite{bandarkar2023belebele}.

Our effort is part of a larger project on Arabic QA; as such we deemed the {\sc Belebele} dataset to be a suitable starting point for establishing an EQA benchmark for Arabic dialect QA when paired with MSA and English passages. 
Our five dialects of focus represent the full spectrum of variations across the Arab world: Gulf, Levantine, Iraqi, Egyptian, and Moroccan. We have questions in English, MSA and Arabic dialects, with the answers in MSA and English only.

Broadly defined, our first task in the conversion from MCQA to EQA is to identify the spans in the passages that answer the MCQs.

\subsection{The MCQ Zoo}
\label{sec:MCQzoo}

When examining the different {\sc Belebele} MCQs and their associated passages, we recognized a considerable number of MCQ types that vary on a continuum from simple word matching to required passage and world understanding. 
Table~\ref{tab:belebele_examples} presents a set of English examples along this continuum to highlight the challenges that hindered us from employing a fully automated process in identifying text spans in the associated passages.  
The Arabic version is in Appendix~\ref{app:ar-examples}. 

\paragraph{}The simplest MCQ types are perhaps \textbf{Exact Match}, Partial \textbf{Overlap} Match and \textbf{Morph}ological Match, Table~\ref{tab:belebele_examples}.(a, b, c), respectively, where the MCQ answer is ``present'' in the passage. 
%
%
%
%
%
%
Relatively more complex are cases that may be resolvable through \textbf{Paraphrasing}, \textbf{Entailment}, or \textbf{Semantic Similarity}, in Table~\ref{tab:belebele_examples}.(d, e, f), respectively.
Next comes the cases that require higher levels of awareness of  context, from that of \textbf{Coreference} in the text all the way to \textbf{Pragmatic Knowledge}, in Table~\ref{tab:belebele_examples}.(g-i).
%
%
%
%
%
%
%
Finally we present examples of complex kinds of MCQ, where wording of the question is highly dependent on the provided multi-choice answers. While a span can be identified in principle in Table~\ref{tab:belebele_examples}.(j), the question may not stand alone with the passage in Table~\ref{tab:belebele_examples}.(k, l).


\paragraph{}These challenges made it apparent that (a) not all MCQAs are usable for the EQA task; and (b) that human annotation was a more realistic approach than automating the conversion of the dataset.  While the first few MCQ types suggest that an automatic process could be used to identify the QA span, the rest, in increasingly difficulty, show that the MCQ answers are not as helpful. The last two types are not usable in our assessment for this task.  We took all of these insights into consideration as we developed the guidelines and annotated the spans,  which we discuss next.

\section{Annotation Process}
\label{sec:annotation}
In this study, \textit{annotation} refers to marking the appropriate `answer span' in the original passage in response to a question, or labelling a question as unanswerable (`X'), as shown in Table~\ref{tab:belebele_examples}.  Our annotations focused on EN and MSA parallel passages from the {\sc Belebele} dataset and we call our annotated set {\sc Belebele-EQA}. Our repurposing approach involved the following steps. 


\subsection{Automatic Filtering}

The original {\sc Belebele} dataset contains 900 question-answer pairs. However, as highlighted in Section~\ref{sec:MCQzoo}, many of these multiple-choice questions did not suit the EQA task. As a first pass, we automatically removed QA pairs that were deemed unsuitable based on the following set of keywords: ``\textit{Which of the following}'', ``\textit{Select all that apply}", ``\textit{Choose the correct}'', ``\textit{According to the passage}'', ``\textit{Based on the information given}''. 
Given that the dataset is parallel, we carried out this automatic filtering on the English content first and subsequently removed the same QA pairs in the corresponding MSA content. After this first filtering pass, we were left with 415 QA pairs, whose EQA answer spans we annotated manually.

\subsection{Pilot study} We first carried out a pilot annotation study in order to fully understand the nature of the {\sc Belebele} dataset and develop our annotation guidelines. The pilot annotation study was carried out on the first 100 QAs for both EN and MSA. The annotators were presented with the {\sc Belebele}  passage, the original {\sc Belebele}  question, the `correct answer' from the multiple-choice answer list\footnote{Available from the {\sc Belebele} data release \url{https://github.com/facebookresearch/belebele}} 
and a column in which to write the span. 
Two native Arabic speakers annotated 50 MSA QAs each, and two fluent English speakers each annotated 50 EN QAs. 
It was during this pilot study that the extent of the challenges highlighted in Section~\ref{sec:challenges} became known. 

\subsection{Annotation Guidelines} Informed by the findings of the initial pilot study, we created two separate versions of the guidelines for English and Arabic, accounting for linguistic differences that influenced the wording of the instructions and the need for language-specific examples. See Appendices~\ref{app:guidelines-en} and~\ref{app:guidelines-ar}  for the complete EN and MSA guidelines, respectively.

Also informed by the pilot study,  the original multiple choice answer was hidden during the annotation, as it had been deemed to be more distracting than helpful in the pilot round (see discrepancies between MC answers and spans in Table~\ref{tab:belebele_examples}).

As part of the annotation task, the annotators would mark the span in a copy of the passage using \$!(\texttt{text})!\$ as a span delimiter, while keeping the span as short as possible -- in the style of SQUAD~\cite{rajpurkar-etal-2016-squad}. 

Annotators were also given clear guidance on what to do with articles, prepositions, punctuation, and instances where the answer appeared more than once in the passage (Appendices~\ref{app:guidelines-en} and~\ref{app:guidelines-ar}). Equipped with better awareness of the issues described in Section~\ref{sec:MCQzoo}, for each QA pair, the annotators could either: (i) annotate the span and if necessary, label the QA as `BB' to indicate it as being complex and problematic, as per Table~\ref{tab:belebele_examples}.(d,e,g,i) or (ii) label the QA as `X' to be removed, due to the answer being impossible to find, as per Table~\ref{tab:belebele_examples}.(k,l). 
Finally, annotators were advised that it was more important to find a relevant span (in which evidence of the answer could be found) than finding an exact span that may not exist. This guidance was a significant help given the many differences between the nature of MCQA and EQA. 

\subsection{English Data Annotation\label{subsec:EnDataAnno}} We took a bootstrapping approach to annotation across the two languages. Given the parallel nature of the dataset, the EN annotation was carried out first, followed by a semi-automatic approach to annotating the MSA data (Section~\ref{subsec:ArDataAnno}). The annotations were done on the remaining 415 QA pairs in the {\sc Belebele} dataset (including a review and correction of the first 100 from the pilot round). Eleven annotators, both fluent and native English speakers, annotated the QA pairs, with the annotations subsequently reviewed by independent reviewers.

The distribution of the QA pairs was as follows: 50 QA pairs were used in our IAA study (see Section~\ref{sec:IAA}) and were annotated by three annotators. The remaining 365 pairs were distributed amongst the other eight annotators, where six annotators had 50 pairs each, and the remaining two annotators had 42 and 23, respectively. 
The annotators were given the English {\sc Belebele} passage, the original 
question and a copy of the passage
to mark the answer span.
They could also add 
comments for problematic instances that required broader discussion or verification.

As a result of the annotation process, 86 out of the 415 QA pairs were labelled as `X' and excluded. Out of the remaining 329 pairs, 44 were labelled as `BB'. These BB QAs remain part of the final dataset, as we are also interested in establishing what difficulties they pose for a QA system, given the need for reasoning and resolving linguistic phenomena like paraphrasing, pragmatics and coreference resolution.

\subsection{Arabic Data Annotation\label{subsec:ArDataAnno}} For the MSA data, we exploited the parallel nature of the dataset to project pre-annotations from the EN passages to the MSA passages as part of a bootstrapping approach. The goal was to maximize the alignment between the EN and the MSA spans that provide answers to the questions. Our semi-automated approach followed the following steps:
    \begin{enumerate}
    \vspace{-5pt}
\setlength{\itemsep}{0pt}
\setlength{\parsep}{0pt}
\setlength{\topskip}{0pt}
     \item Marking as `X' the MSA questions labelled `X' in English to avoid re-annotation. In our experience, this approach proved consistent for both English and MSA.
    \item Translation of the EN span to MSA using Helsinki MT tool~\cite{tiedemann-thottingal-2020-opus} 
    \item Using a sliding-window approach to identify the MSA span with the same length and highest n-gram overlap with the EN span.
    \item Marking an approximate location of the EN span in the MSA passage
    \item  If no equivalent translation was found, an approximate MSA location based on the EN span's word offset was used.\footnote{An example of this case is when the  span is 1-2 words long and its translation is not found in the MSA passage.} 
    \item Native Arabic speakers reviewed the pre-annotated spans according to the MSA-based annotation guidelines and made corrections where necessary. 
    \end{enumerate}

The 415 QA pairs were divided amongst four native Arabic speaking annotators, with the annotations subsequently reviewed by independent reviewers. The annotators were presented with the MSA {\sc Belebele} passage, the original MSA {\sc Belebele} question and a copy of the passage where the span was \textit{pre-annotated}. During the review process, the annotators could either: leave the span unchanged or \textit{modify} the span markers, and label as `BB' where necessary, or label the QA as `X' to be removed.

The MSA guidelines are generally similar to the EN guidelines except for containing Arabic examples instead of English. The MSA guidelines match EN in using white-space boundaries as the primary word delimiter. This unavoidably leads to a difference in span scope from EN since Arabic is a morphologically rich language with numerous clitics. As a result some  proclitic prepositions, conjunctions, and the definite article, as well as pronominal enclitics are kept inside the spans.

    
    
\subsection{Inter-Annotator Agreement} \label{sec:IAA}
We carried out an inter-annotation (IAA) study in order to assess the level of agreement across annotators, and get an indication of how reliable the guidelines were.\footnote{As the dataset was small and fully manually reviewed, the IAA study did not serve to evaluate the data quality.} For the IAA study, we selected 50 English QA pairs for annotation by three separate annotators. The three annotations for each QA pair were then compared. 
Instead of calculating an F-score measure, which would give indication of quality of annotations, we utilize the $\gamma$ measure for Inter-Annotator Agreement and Alignment~\cite{Gamma:2015:IAA}.\footnote{Gamma ranges between 0 and 1 where a higher value indicates higher agreement between annotators.} This measure involves identifying the optimal alignment among annotations provided by multiple annotators, aiming to align annotations with high similarity and compute the average dissimilarity between them. Specifically, the average dissimilarity is calculated by averaging dissimilarities among aligned annotations, considering both their categories and positions. The alignment with the minimum mean dissimilarity, selected from all potential alignments initially evaluated during computation, is employed. The final inter-annotator agreement~($\gamma$ measure) was computed on the three sets of 50 QA pairs yielding a high $\gamma$ value of $0.81$, indicating that the designed annotation guidelines are reliable.
%
Given that the MSA annotation guidelines were based on the EN guidelines, we posit that their quality was just as high. The three sets of 50 EN IAA QAs were merged and reviewed to produce a final version for inclusion in our parallel dataset. See Appendix~\ref{app:IAA} for more details and visualization of the span combination process.

\section{Experiments and Results~\label{sec:experiments}}


\subsection{Experimental Setup}

\paragraph{Datasets} We use two datasets in our experiments: the commonly used state-of-the art QA dataset, SQuAD~\cite{rajpurkar-etal-2016-squad} and our {\sc Belebele-EQA}, both of which are Wikipedia-based. For SQuAD, we only used the validation portion of the dataset which contains 10,657 English EQA pairs. 
For {\sc Belebele-EQA}, we have two versions: (i) {\sc Belebele-EQA-}\textbf{All} contains 329 QA pairs for EN, MSA and five dialects, and (ii) {\sc Belebele-EQA-}\textbf{Sub} excludes 44 BB-annotated questions, leaving 285 QA pairs. 



\paragraph{QA Evaluation Tool} To run the QA task on the above datasets, we used PrimeQA~\cite{sil2023primeqa}, an open-source\footnote{\url{https://github.com/primeqa/primeqa}} tool that is built on top of the HuggingFace~\cite{wolf2020huggingfaces} library to support research on running different QA tasks. 
We used two existing models, namely \textbf{PrimeQA/NQ+TyDi}\footnote{PrimeQA/nq\_tydi-reader-xlmr\_large-20221210} and \textbf{PrimeQA/NQ+TyDi+SQuAD},\footnote{PrimeQA/nq\_tydi\_sq1-reader-xlmr\_large-20221110} which are 550m parameters XLM-R$_{Large}$~\cite{conneau-etal-2020-unsupervised} based models that were finetuned for the multilingual TyDi QA task~\cite{clark-etal-2020-tydi}. Both models were finetuned on the Natural Questions dataset~\cite{NQdata} and TyDi~\cite{clark-etal-2020-tydi}. The latter, however, was further fine-tuned on the training portion of the SQuAD dataset. 

\paragraph{Evaluation Metrics} We report the results using the commonly used  F1-score and  Exact Match (EM) score.
F1-score is the harmonic mean of Precision and the Recall on  word-level uni-grams. The default F1 and EM scores as implemented in PrimeQA toolkit preprocess references and predictions by dropping punctuation, extra white space, and for EN, lower casing and removing the articles \textit{a/an/the}.
We also introduce two normalized variants, F1$^n$ and EM$^n$, where we normalize references and predictions by removing the respective language stopwords as defined in NLTK \cite{bird-loper-2004-nltk}.


\subsection{Results and Analysis}\label{resultsanalysis}
In this section, we present the evaluation results for the {\sc Belebele-EQA} and SQuAD datasets. 

\paragraph{SQuAD vs. BELEBELE-EQA} 
We evaluate the SQuAD dataset as well as the EN-EN QA pairs from the {\sc Belebele-EQA} dataset using the PrimeQA toolkit (See Appendix~\ref{PrimeQA} for details)
%
This experiment allows us to (a) verify that the chosen pretrained models are capable of solving the QA task at-hand without further finetuning, and (b) establish a sense of \textit{difficulty} for the {\sc Belebele-EQA} datasets compared to the commonly used SQuAD. 
%
%
We observe the following based on the results in Table~\ref{tbl:baseline}:
First, on SQuAD, the F1 shows a negligible difference between the two pretrained models, while EM is surprisingly lower by 3.4\% for the model that was finetuned on the SQuAD dataset.\footnote{The state-of-the-art F1 and EM scores on  SQuAD are 93.2 and 90.9, respectively, as per the public PapersWithCode leaderboard: \url{https://paperswithcode.com/sota/question-answering-on-squad20}}
In contrast, the results on {\sc Belebele-EQA-}All show that additional finetuning on SQuAD yielded better performance both in F1 and EM.

Second, the best {\sc Belebele-EQA} (All and Sub) F1 scores (71.0 and 76.6) are 18.6 and 14 points less than SQuAD's F1 score, respectively. Similarly, the best respective EM scores are 50.5 and 56.1, or 28.6 and 23 points less than that of the SQuAD dataset. 
Finally, and as expected, removing the {\sc Belebele-EQA} hard questions ({\sc Belebele-EQA-}Sub) resulted in better performance on both F1 and EM measures. 

%
%

\paragraph{BELEBELE-EQA Results} 
Based on the baseline results presented above, we report in Table~\ref{tab:Results} on the different question language setups using the best identified model: {\textbf{PrimeQA/NQ+TyDi+SQuAD}}.
For specific experiments, we refer to the pair of languages in \textbf{passage}-\textbf{question}, e.g., EN-EN, or MSA-Iraqi.

\paragraph{\textit{Monolingual Experiments (EN-EN, MSA-MSA)}} 

        
The basic standard monolingual setups have the highest performing F-1 scores, with EN being higher than MSA by 9\% absolute.  Despite being a multilingual model, this performance difference is not unexpected given the relative resource richness of EN compared with MSA.

\begin{table}[t!]
\small
    \centering
    \setlength{\tabcolsep}{8pt}
    \begin{tabular}{l|r|c|c}
    \hline
    \textbf{Dataset} & \textbf{Questions} & \textbf{F1}  & \textbf{EM} \\\hline
    \multicolumn{4}{c}{\textbf{PrimeQA/NQ+TyDi}} \\\hline
    SQuAD & 10,570 & 90.5 & 82.5  \\
    {\sc Belebele-EQA}-All & 329 & 68.1 & 46.5 \\ 
    {\sc Belebele-EQA}-Sub & 285 & 71.3 & 50.5\\ \hline
    
    \multicolumn{4}{c}{\textbf{PrimeQA/NQ+TyDi+SQuAD}} \\\hline
    SQuAD & 10,570 & 90.6 & 79.1   \\
    {\sc Belebele-EQA}-All & 329 & 71.0 & 50.5 \\ 
    {\sc Belebele-EQA}-Sub & 285 & 76.6 & 56.1 \\ \hline

    \end{tabular}
    \caption{Evaluating the SQuAD and {\sc belebele-EQA} datasets using the PrimeQA dataset. (EM: Exact Match. Scores are percentages. Higher is better.)~\label{tbl:baseline}}
\end{table}

\begin{table*}[t!]
\small 
    \centering
   \setlength{\tabcolsep}{10pt}
    \begin{tabular}{@{}l@{ }|l||cc|cc||cc|cc@{ }}
       \hline                    
\multirow{2}*{\textbf{Passage}} & \multirow{2}*{\textbf{Question}} & \multicolumn{4}{c||}{\textbf{All}}   &  \multicolumn{4}{c}{\textbf{Sub}}   \\ 
 &  & \textbf{F1} & \textbf{EM} & \textbf{F1$^n$} & \textbf{EM$^n$} & \textbf{F1} & \textbf{EM} & \textbf{F1$^n$} & \textbf{EM$^n$} \\ \hline\hline 
    \multirow{14}{*}{EN} &EN &71.0 &50.5 &74.0 &60.2 &76.6 &56.1 &79.5 &65.6 \\
    &MSA &59.9 &39.8 &62.7 &47.7 &62.9 &43.5 &65.6 &51.2 \\ \cdashline{2-10}
    &DA--Iraqi &44.5 &27.7 &46.2 &31.3 &46.0 &29.5 &47.8 &33.0 \\
    &DA--Levantine &49.5 &31.9 &51.2 &36.5 &51.9 &35.1 &53.7 &39.6 \\
    &DA--Gulf &50.4 &31.3 &52.6 &37.4 &52.1 &33.0 &54.2 &38.9 \\
    &DA--Morrocan &36.3 &18.5 &38.5 &21.0 &37.0 &20.0 &39.3 &22.1 \\
    &DA--Egyptian &48.8 &30.1 &51.5 &36.2 &51.3 &33.3 &54.0 &39.3 \\
    &\textit{DA--Average} & 
45.9 & 27.9 & 48.0 & 32.5 & 47.7 & 30.2 & 49.8 & 34.6\\ \cdashline{2-10}
    &DA--Iraqi (T$_{EN}$) &59.8 &39.2 &62.7 &46.8 &64.2 &42.8 &67.2 &50.9 \\
    &DA--Levantine (T$_{EN}$) &65.4 &46.2 &68.2 &53.5 &69.7 &50.2 &72.3 &57.9 \\
    &DA--Gulf (T$_{EN}$) &65.4 &43.8 &68.2 &52.9 &69.5 &48.1 &72.2 &57.2 \\
    &DA--Morrocan (T$_{EN}$) &63.0 &41.0 &65.7 &49.5 &67.0 &44.9 &69.6 &53.0 \\
    &DA--Egyptian (T$_{EN}$) &65.7 &45.0 &68.7 &53.8 &70.0 &49.5 &72.9 &58.2 \\
    &\textit{DA--Average} (T$_{EN}$) & 
    63.9 & 43.0 & 66.7 & 51.3 & 68.1 & 47.1 & 70.8 & 55.4 \\\hline\hline

    \multirow{14}{*}{MSA} &EN &60.4 &41.6 &62.3 &46.8 &64.5 &46.0 &66.7 &50.9 \\
    &MSA &62.0 &40.7 &64.3 &46.2 &65.6 &44.2 &68.1 &49.5 \\\cdashline{2-10}
    &DA--Iraqi &50.3 &32.2 &51.2 &35.0 &54.2 &36.1 &55.6 &39.3 \\
    &DA--Levantine &54.2 &35.3 &55.7 &38.6 &58.5 &39.3 &60.4 &43.2 \\
    &DA--Gulf &56.6 &36.2 &58.4 &40.7 &59.3 &39.6 &61.3 &44.2 \\
    &DA--Morrocan &45.1 &26.7 &46.6 &30.1 &48.1 &29.8 &49.6 &33.3 \\
    &DA--Egyptian &54.5 &34.0 &56.6 &39.2 &57.3 &37.5 &59.5 &42.5 \\
    &\textit{DA--Average} &  
52.1 & 32.9 & 53.7 & 36.7 & 55.5 & 36.5 & 57.3 & 40.5\\\cdashline{2-10}

    &DA--Iraqi (T$_{MSA}$) &51.5 &31.3 &53.1 &35.3 &53.9 &34.0 &55.7 &37.9 \\
    &DA--Levantine (T$_{MSA}$) &58.5 &38.3 &60.3 &43.2 &62.1 &42.1 &64.1 &46.7 \\
    &DA--Gulf (T$_{MSA}$) &57.9 &38.0 &59.9 &42.9 &61.0 &40.7 &63.3 &46.0 \\
    &DA--Morrocan (T$_{MSA}$) &56.5 &36.2 &58.6 &42.6 &59.4 &39.3 &61.5 &45.3 \\
    &DA--Egyptian (T$_{MSA}$) &58.5 &37.7 &60.4 &42.2 &62.1 &41.1 &64.2 &46.0 \\
    &\textit{DA--Average} (T$_{MSA}$) & 
    56.6 & 36.3 & 58.5 & 41.2 & 59.7 & 39.4 & 61.8 & 44.4 \\ \hline
    
    \end{tabular}
    \caption{Monolingual and Cross-lingual evaluation of the {\sc Belebele-EQA} dataset using the PrimeQA toolkit with PrimeQA/NQ+TyDi+SQuAD. (EM: Exact Match; F1$^n$ and EM$^n$ are the text-normalized versions of F1 and EM, respectively; DA: Dialectal Arabic; T$_{EN/MSA}$: Question translated to EN/MSA; Scores are \%s; Higher is better.)}
    \label{tab:Results}
\end{table*}

\paragraph{\textit{Cross-Lingual Experiments (MSA-EN, EN-MSA)}} 
The basic cross-lingual setups  are next in rank of performance with MSA-EN F1 being only slightly better than EN-MSA (0.5\% absolute).  The performance drop due to question language is much higher for EN passages (11.1\% absolute) compared to MSA passages (1.6\% absolute). This suggests that the model struggles in cross-lingual settings in ways independent of its monolingual performance.

\paragraph{\textit{Dialectal Experiments (MSA-DA, EN-DA)}} 

The DA questions consistently lead to lower performance compare to the above-mentioned settings, and with performance with MSA passages being better on average than on EN passages (by 6.3\% absolute).  The specific dialects seem to follow a common order of Gulf$>$(Egyptian|Levantine)$>$Iraqi$>$Moroccan, which we speculate may reflect the order of their question sets' similarity to MSA. 
%
Automatically translating dialectal questions into the target passage language using Google Translate\footnote{For DA-to-English, we translated using Google Translate (AR-to-EN), while for DA-to-MSA, we used English as an intermediate language (DA $\Rightarrow$ EN  $\Rightarrow$  MSA): Google Translate (AR-to-EN  $\Rightarrow$  EN-to-AR).} results in notable improvements: an average increase of 18\% for English passages and 4.4\% for MSA passages (All setting). 
Moroccan saw the greatest benefit from this translation method, with F1 increases of 26.7\% for EN and 11.4\% for MSA.

\paragraph{\textit{BELEBELE-EQA  All vs Sub}} 

Across all experiments, the easier subset (Sub) has higher scores than (All): on average, 2.6\% absolute F1 increase for EN passages, and 1.5\% for MSA passages. The highest difference is for EN-EN (5.4\% absolute F1) and the lowest for MSA-Gulf (0.6\%).							

\paragraph{\textit{Normalized F1 and EM}} 
The normalization results are generally consistent in rank order; but the effect of normalization in EN is much higher than MSA: F1 (2.3\%  vs 1.7\%) and EM (5.8\% 4.3\%).  
We note that the normalization process reduces the length of references and predictions by 30\% for EN, but only 10\% for MSA.

\section{Discussion}
\label{sec:discussion}

Our motivation for this study was to create a practical and reliable multilingual EQA benchmark dataset, without the need to start from scratch or resort to creating synthetic data. Given our languages of interest, repurposing the {\sc Belebele} MCQA dataset seemed like a feasible solution. However, answering the research question on how feasible this approach is involves considering the various insights and findings presented in this paper. 

We found that while repurposing is possible, 
our exploratory study showed that it required significant manual effort. Contrary to initial assumptions, we discovered that the original answers from the multiple-choice question pairs were not suitable for automated approaches. In fact, the only multiple-choice types in Table~\ref{tab:belebele_examples} that resemble non-complex SQUAD-style QA pairs are \textit{(a) Exact}. 

Rather than offering a scalable solution here, instead our manual efforts and the subsequent insights serve to make this repurposing task much more feasible and potentially automated by other dataset developers. 
We can see that the repurposing process is much more feasible when the differences between the QA styles are fully understood. We found that, through an effective filtering process, we could easily eliminate many QA pairs that did not lend themselves easily to EQA. While this resulted in reducing the dataset size by more than half, this leaves open new avenues for examining how those `unsuitable' QA pairs could be reformulated to be useful in EQA. 
%
Repurposing a multilingual dataset like {\sc Belebele} means that its parallelism could be exploited through a bootstrapping approach. We have learned that once one language dataset is manually annotated and verified, the spans could be relatively easily projected to a new language from the wider dataset provided a translation system is available (see Section~\ref{subsec:ArDataAnno}). 
Likewise, few changes were required when adapting our annotation guidelines from EN to MSA, suggesting that this adaptability may hold for other languages.
Finally, our experimental results confirm that our approach produced a useful data set that can be used for benchmarking cross-lingual EQA for all {\sc Belebele} languages. 

\section{Conclusion and Future Work}
\label{sec:conclusion}

We repurposed the MCQA {\sc Belebele} dataset to create a new EQA dataset for English and MSA, with cross-lingual benchmark results for dialectal Arabic QA.\footnote{\url{https://github.com/mbzuai-nlp/MultiChoice2ExtractiveQA}}
We also provided guidelines for practitioners exploring QA in other {\sc Belebele} languages. Our work, including negative insights, aims to enhance understanding of the challenges and opportunities in reformulating NLP tasks.



For future work, we plan to support repurposing {\sc Belebele-EQA} for other languages, potentially automating the process. The dataset offers a resource for evaluating such methods. We also aim to use it for fine-tuning general-purpose Arabic QA models and improving EQA evaluation metrics, focusing on span length.

\section*{Limitations}
We would like to point to some limitations of our study.
First, while we are addressing a general problem, in our study, out of all the other {\sc Belebele} under-resourced languages, we only focused on cross-lingual evaluation of five Arabic dialects. While we believe that our findings are more generally applicable, we would like to note that each language and dataset may present unique challenges and opportunities.
Second, our repurposing was based on a single dataset, {\sc Belebele}, and for a single task (extractive question answering in the style of machine reading comprehension), and we note that other tasks may require some specific considerations. 

\section*{Ethics and Broader Impact}

\paragraph{Data Bias.} It is important to acknowledge the possibility of biases within our dataset due to subjective human judgments, or due to the influence of the original task's data (e.g. a narrow cultural representation across the questions).

\paragraph{Broader Impact and Potential Use.} Our repurposing insights presented here can help speed up the creation of resources and tools for resource-poor languages and dialects. We do not see any immediate threats from its use.

\bibliography{custom}



\clearpage
\onecolumn
\appendix 
\section{Processing IAA Spans for Visualization}\label{app:IAA}

\begin{figure*}[h!]
    \centering
    \includegraphics[width=\textwidth]{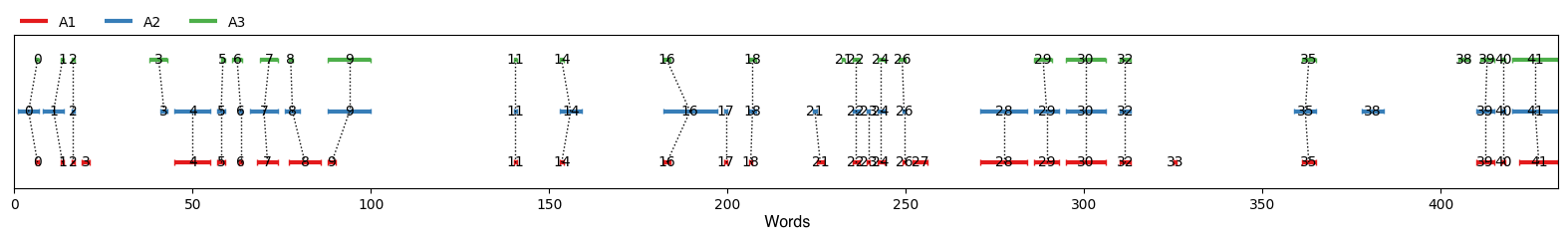}
    \caption{Visualization of the IAA agreement for 42 questions. The colored lines indicate a span. Each color represents an annotator. The number on the colored line is a question ID.}
    \label{fig:IIA}
\end{figure*}

In Figure~\ref{fig:IIA}, we visualize the degree of agreement among the annotators on the 42 QA pairs that remained from the IAA study (See Section~\ref{sec:IAA}). We combine the annotation spans into one artificial document in order to visualize the results in a meaningful way where this process is explained later in this section. 

In this figure, the x-axis represents the artificial document that is 450 words long. The vertical, colored lines represents a span in that document. The three different colors represent the three annotators and the number on the span is the question number. 42 out of the 50 QA pairs remained as eight questions were labelled as `X' by all three annotators and were excluded. 
As annotators were allowed to label a questions as `X', some questions have less than three annotations, e.g. Question 38.

\subsection{Combining the Spans Into One Artificial Document.}
While we could visualize the agreement between annotators for each QA pairs separately, this would have resulted in 42 different figures, one for each QA pair. Instead, we decided to combine the annotated spans from each annotator to create one large artificial document where the annotated spans would appear in different locations. One simple approach to this is combining the whole passages which will offset the location of the annotated span by the length of all the passages that came before it. This, however, will result in a huge document that is too long to have a meaningful visualization. 

\subsection{Compressing the Artificial Document.}
To overcome this issue, we \textit{compressed} the document by removing the large un-annotated text spans. Note that the annotation are technically a start and end positions in a document and our task is basically to measure (and visualize) the chosen start and end position of each QA pair across the three annotators. This means that the QA pairs are independent, and ordering them in the document or shifting their locations --as long as they do not overlap and they maintain their original length-- should not affect them. 

To illustrate, assume that annotator ($a_1$) for passage ($p_1$) chose the span (3, 7) indicating that the answer is from word ($w_3$) to ($w_7$) while annotator ($a_2$) chose the span (4, 10). We can see that ($a_1$) and ($a_2$) \textit{disagree} in the regions: ($w_4$) and from ($w_8$) to ($w_10$). If we add passage ($p_0$) which contains 20 words before ($p_1$) to make one document resulting in shifting all the words in ($p_1$), as well as in the spans ($a_1$) and ($a_2$) by 20 words, we see that the disagreement remains the same. After the shift, ($a_1$) and ($a_2$) change from (3, 7) and (4, 10) to (23, 27) and (24, 30), respectively. As can be noticed, the disagreement between ($a_1$) and ($a_2$) remained in two regions with the same length for each region, that is ($w_24$) and ($w_28$) to ($w_30$).

\section{PrimeQA Experimental Details}\label{PrimeQA}
The experiments discusses in Section~\ref{resultsanalysis} were conducted on an Apple M1 Max MacBook Pro with 32GBs of memory. Since we used the aforementioned models in inference mode, each experiment with 329 questions (E.g. the combination of En Passage and En Questions) took approximately 30 seconds to complete. Note that additional time may be required to download each model for the first time. The 56 core experiments (without translation) in Table~\ref{tab:Results} (2 Passages x 7 question types x 2 configurations x 2 normalization schemes) are expected to require 28 minutes of running.

\newpage

\section{Arabic Examples of MCQ-Span Types}
\label{app:ar-examples}
\begin{table*}[h!]
    \centering
    \includegraphics[width=\textwidth]{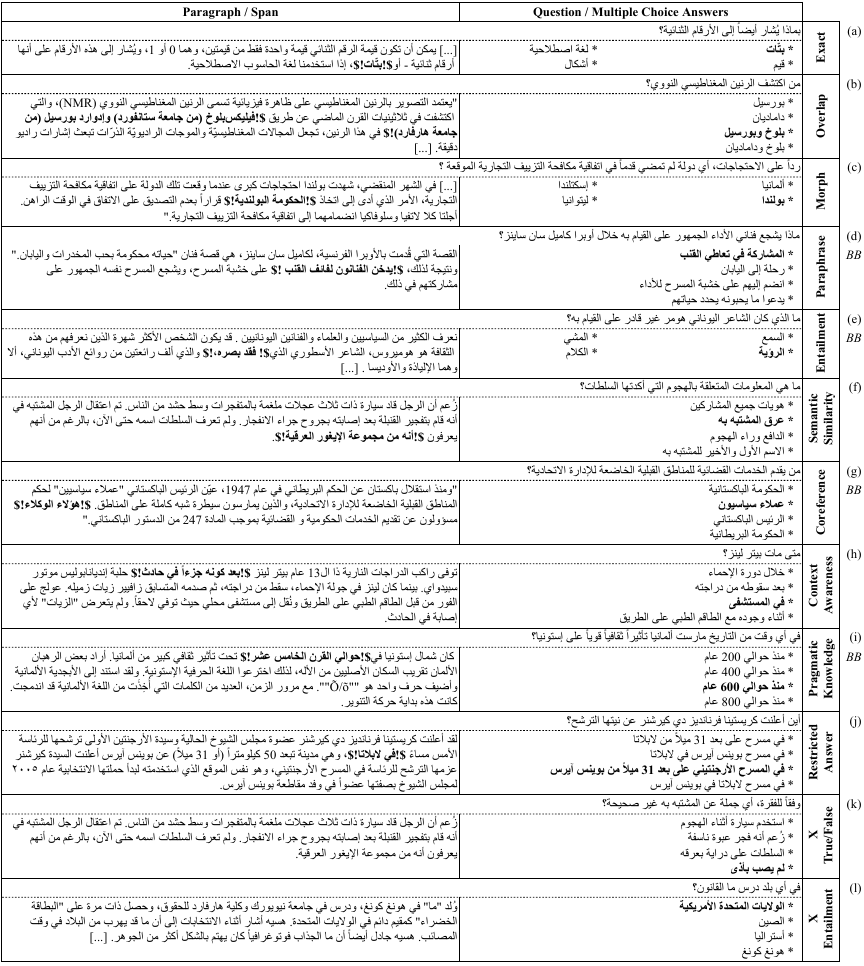}
    \caption{This table shows examples in Arabic demonstrating part of the range of types of MCQs in {\sc Belebele}, and the respective issues they pose for extractive QA in this work. These examples are parallel to the English examples in Table~\ref{tab:belebele_examples}.}
    \label{tab:belebele_examples_ar}
\end{table*}
\vspace{3cm}

\section{Guidelines for English Answer Span Annotation}
\label{app:guidelines-en}
\vspace{0.8in}
\begin{center}

\includegraphics[trim={3cm 3cm 3cm 3cm},scale=0.9]{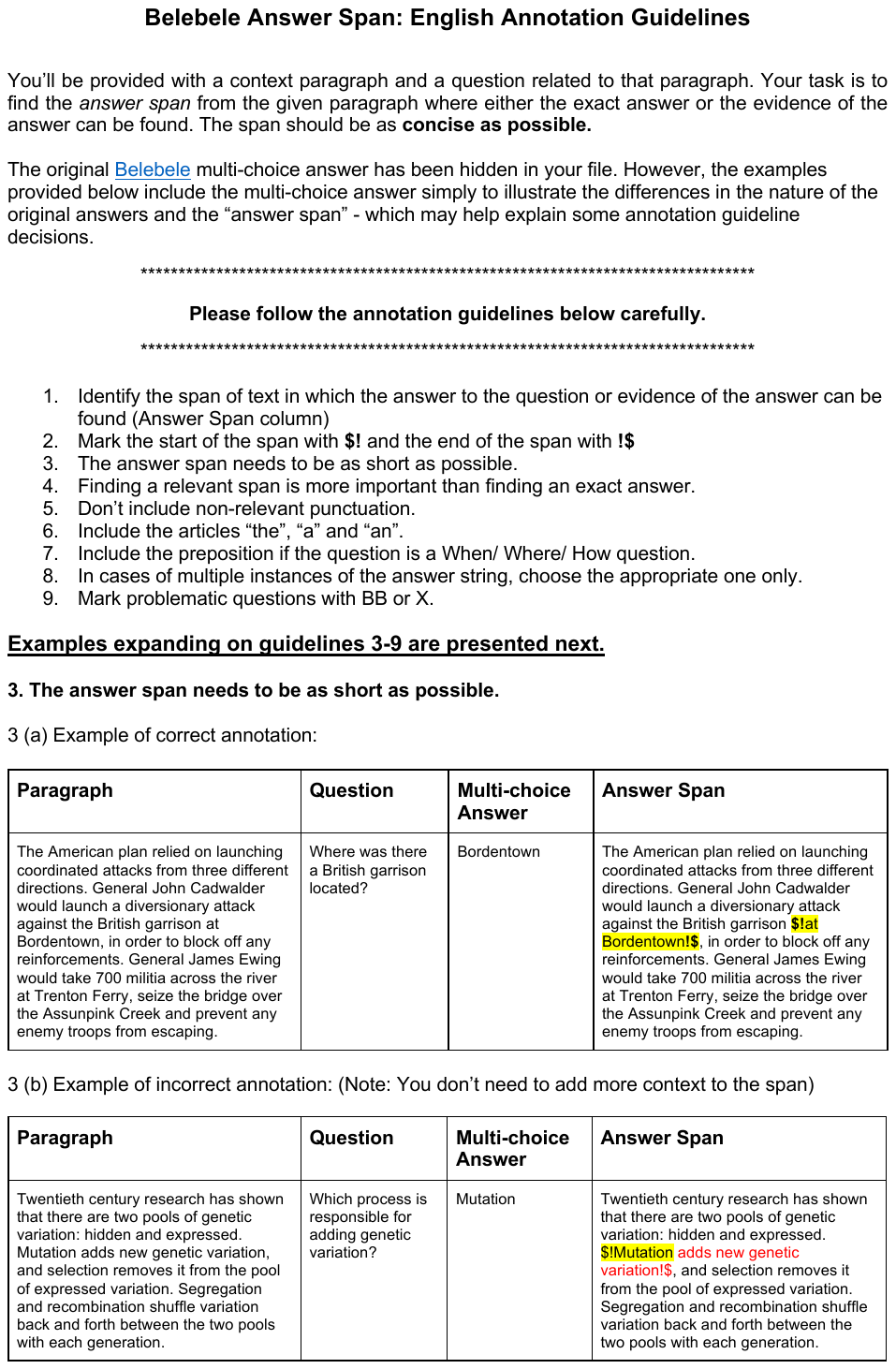}

\includepdf[pages=2-6,scale=0.9]{BelebeleQA-Annotation-Guidelines_EN_Paper_Appendix.pdf}

\section{Guidelines for Modern Standard Arabic Answer Span Annotation}
\label{app:guidelines-ar}
\vspace{0.8in}
\includegraphics[trim={3cm 3cm 3cm 3cm},scale=0.9]{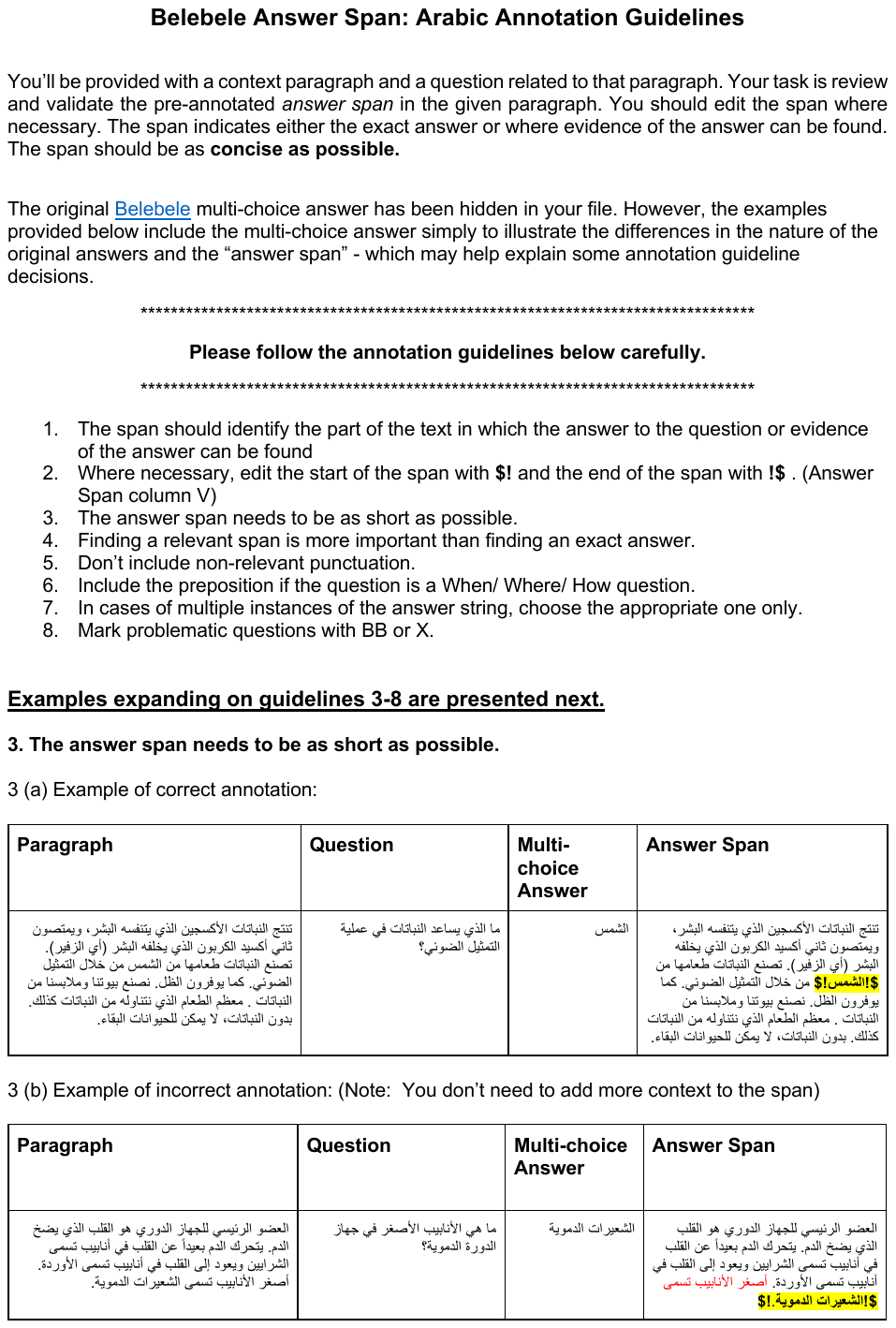}



\section*{Supplementary material}

# B Arabic Examples of MCQ-Span Types

| Paragraph / Span | Question / Multiple Choice Answers | | |
|---|---|---|---|
| [...] يمكن أن تكون قيمة الرقم الثنائي قيمة واحدة فقط من قيمتين، وهما 0 أو 1، ويُشار إلى هذه الأرقام على أنها أرقام ثنائية - أو **$!بتّات!$**، إذا استخدمنا لغة الحاسوب الاصطلاحية. | بماذا يُشار أيضاً إلى الأرقام الثنائية؟<br>* **بتّات**       * لغة اصطلاحية<br>* قيم          * أشكال | Exact | (a) |
| "يعتمد التصوير بالرنين المغناطيسي على ظاهرة فيزيائية تسمى الرنين المغناطيسي النووي (NMR)، والتي اكتشفت في ثلاثينيات القرن الماضي عن طريق **$!فيليكس بلوخ (من جامعة ستانفورد) وإدوارد بورسيل (من جامعة هارفارد)!$** في هذا الرنين، تجعل المجالات المغناطيسيّة والموجات الراديويّة الذرّات تبعث إشارات راديو دقيقة. [...] | من اكتشف الرنين المغناطيسي النووي؟<br>* بورسيل          * داماديان<br>* **بلوخ وبورسيل**<br>* بلوخ وداماديان | Overlap | (b) |
| [...] في الشهر المنقضي، شهدت بولندا احتجاجات كبرى عندما وقعت تلك الدولة على اتفاقية مكافحة التزييف التجارية. الأمر الذي أدى إلى اتخاذ **$!الحكومة البولندية!$** قراراً بعدم التصديق على الاتفاق في الوقت الراهن. أجلتا كلا من لاتفيا وسلوفاكيا انضمامها إلى اتفاقية مكافحة التزييف التجارية." | رداً على الاحتجاجات، أي دولة لم تمضي قدماً في اتفاقية مكافحة التزييف التجارية الموقعة ؟<br>* ألمانيا       * إسكتلندا<br>* **بولندا**       * ليتوانيا | Morph | (c) |
| القصة التي قُدمت بالأوبرا الفرنسية، لكاميل سان سايرس، هي قصة فنان "حياته محكومة بحب المخدرات واليابان." ونتيجة لذلك، **$!يدخن الفنانون لفائف القنب !$** على خشبة المسرح، ويشجع المسرح نفسه الجمهور على مشاركتهم في ذلك. | ماذا يشجع فناني الأداء الجمهور على القيام به خلال أوبرا كاميل سان سايرس؟<br>* **المشاركة في تعاطي القنب**<br>* رحلة إلى اليابان<br>* انضم إليهم على خشبة المسرح للأداء<br>* يدعوا ما يحبونه يحدد حياتهم | Paraphrase<br>BB | (d) |
| نعرف الكثير من السياسيين والعلماء والفنانين اليونانيين . قد يكون الشخص الأكثر شهرة الذين نعرفهم من هذه الثقافة هو هوميروس ، الشاعر الأسطوري الذي **$! فقد بصره،!$** والذي ألف رائعتين من روائع الأدب اليوناني، ألا وهما الإلياذة والأوديسا . [...] | ما الذي كان الشاعر اليوناني هومر غير قادر على القيام به؟<br>* السمع         * المشي<br>* **الرؤية**       * الكلام | Entailment<br>BB | (e) |
| زعم أن الرجل قاد سيارة ذات ثلاث عجلات ملغمة بالمتفجرات وسط حشد من الناس. تم اعتقال الرجل المشتبه في أنه قام بتفجير القنبلة بعد إصابته بجروح جراء الانفجار. ولم تعرف السلطات اسمه حتى الآن، بالرغم من أنهم يعرفون **$!أنه من مجموعة الإيغور العرقية!$**. | ما هي المعلومات المتعلقة بالهجوم التي أكدتها السلطات؟<br>* هويات جميع المشاركين<br>* **عرق المشتبه به**<br>* الدافع وراء الهجوم<br>* الاسم الأول والأخير للمشتبه به | Semantic<br>Similarity | (f) |
| "ومنذ استقلال باكستان عن الحكم البريطاني في عام 1947، عيّن الرئيس الباكستاني "عملاء سياسيين" لحكم المناطق القبلية الخاضعة للإدارة الاتحادية، والذين يمارسون سيطرة شبه كاملة على المناطق. **$!هؤلاء الوكلاء!$** مسؤولون عن تقديم الخدمات الحكومية و القضائية بموجب المادة 247 من الدستور الباكستاني." | من يقدم الخدمات القضائية للمناطق القبلية الخاضعة للإدارة الاتحادية؟<br>* الحكومة الباكستانية<br>* **عملاء سياسيون**<br>* الرئيس الباكستاني<br>* الحكومة البريطانية | Coreference<br>BB | (g) |
| توفى راكب الدراجات النارية ذا الـ13 عام بيتر لينز **$!بعد كونه جزءاً في حادث!$** حلبة إنديانابوليس موتور سبيدواي. بينما كان لينز في جولة الإحماء، سقط من دراجته، ثم صدمه المتسابق زافير زيات زميله. عولج على الفور من قبل الطاقم الطبي على الطريق ونُقل إلى مستشفى محلي حيث توفي لاحقاً. ولم يتعرض "الزيات" لأي إصابة في الحادث. | متى مات بيتر لينز؟<br>* خلال دورة الإحماء<br>* بعد سقوطه من دراجته<br>* **في المستشفى**<br>* أثناء وجوده مع الطاقم الطبي على الطريق | Context<br>Awareness | (h) |
| كان شمال إستونيا في **$!حوالي القرن الخامس عشر!$** تحت تأثير ثقافي كبير من ألمانيا. أراد بعض الرهبان الألمان تقريب السكان الأصليين من الأله، لذلك اخترعوا اللغة الحرفية الإستونية. ولقد استند إلى الأبجدية الألمانية وأضيف حرف واحد هو "Ö/õ". مع مرور الزمن، العديد من الكلمات التي أُخذت من اللغة الألمانية قد اندمجت كانت هذه بداية حركة التنوير. | في أي وقت من التاريخ مارست ألمانيا تأثيراً ثقافياً قوياً على إستونيا؟<br>* منذ حوالي 200 عام<br>* منذ حوالي 400 عام<br>* **منذ حوالي 600 عام**<br>* منذ حوالي 800 عام | Pragmatic<br>Knowledge<br>BB | (i) |
| لقد أعلنت كريستينا فرنانديز دي كيرشنر عضوة مجلس الشيوخ الحالية وسيدة الأرجنتين الأولى ترشحها للرئاسة الأمس مساءً **$!في لابلاتا!$**، وهي مدينة تبعد 50 كيلومتراً (أو 31 ميلاً) عن بوينس آيرس أعلنت السيدة كيرشنر عزمها الترشح للرئاسة في المسرح الأرجنتيني، وهو نفس الموقع الذي استخدمته لبدأ حملتها الانتخابية عام 2005 لمجلس الشيوخ بصفتها عضواً في وفد مقاطعة بوينس آيرس. | أين أعلنت كريستينا فرناندز دي كيرشنر عن نيتها الترشح؟<br>* في مسرح على بعد 31 ميلاً من لابلاتا<br>* في مسرح بوينس آيرس في لابلاتا<br>* **في المسرح الأرجنتيني على بعد 31 ميلاً من بوينس آيرس**<br>* في مسرح لابلاتا في بوينس آيرس | Restricted<br>Answer | (j) |
| زُعم أن الرجل قاد سيارة ذات ثلاث عجلات ملغمة بالمتفجرات وسط حشد من الناس. تم اعتقال الرجل المشتبه في أنه قام بتفجير القنبلة بعد إصابته بجروح جراء الانفجار. ولم تعرف السلطات اسمه حتى الآن، بالرغم من أنهم يعرفون أنه من مجموعة الإيغور العرقية. | وفقاً للفقرة، أي جملة عن المشتبه به هي غير صحيحة؟<br>* استخدم سيارة أثناء الهجوم<br>* زُعم أنه فجر عبوة ناسفة<br>* السلطات على دراية بعرقه<br>* **لم يصب بأذى** | X<br>True/False | (k) |
| وُلد "ما" في هونغ كونغ، ودرس في جامعة نيويورك وكلية هارفارد للحقوق، وحصل ذات مرة على "البطاقة الخضراء" كمقيم دائم في الولايات المتحدة. هسيه أشار أثناء الانتخابات إلى أن ما قد يهرب من البلاد في وقت المصائب. هسيه جادل أيضاً أن ما الجذاب فوتوغرافياً كان يهتم بالشكل أكثر من الجوهر. [...] | في أي بلد درس ما القانون؟<br>* **الولايات المتحدة الأمريكية**<br>* الصين<br>* أستراليا<br>* هونغ كونغ | X<br>Entailment | (l) |

Table 4: This table shows examples in Arabic demonstrating part of the range of types of MCQs in BELEBELE, and the respective issues they pose for extractive QA in this work. These examples are parallel to the English examples in Table 1.

# C  Guidelines for English Answer Span Annotation

## Belebele Answer Span: English Annotation Guidelines

You'll be provided with a context paragraph and a question related to that paragraph. Your task is to find the *answer span* from the given paragraph where either the exact answer or the evidence of the answer can be found. The span should be as **concise as possible.**

The original Belebele multi-choice answer has been hidden in your file. However, the examples provided below include the multi-choice answer simply to illustrate the differences in the nature of the original answers and the "answer span" - which may help explain some annotation guideline decisions.

*************************************************************************

**Please follow the annotation guidelines below carefully.**

*************************************************************************

1. Identify the span of text in which the answer to the question or evidence of the answer can be found (Answer Span column)
2. Mark the start of the span with **$!** and the end of the span with **!$**
3. The answer span needs to be as short as possible.
4. Finding a relevant span is more important than finding an exact answer.
5. Don't include non-relevant punctuation.
6. Include the articles "the", "a" and "an".
7. Include the preposition if the question is a When/ Where/ How question.
8. In cases of multiple instances of the answer string, choose the appropriate one only.
9. Mark problematic questions with BB or X.

### Examples expanding on guidelines 3-9 are presented next.

**3. The answer span needs to be as short as possible.**

3 (a) Example of correct annotation:

| Paragraph | Question | Multi-choice Answer | Answer Span |
|---|---|---|---|
| The American plan relied on launching coordinated attacks from three different directions. General John Cadwalder would launch a diversionary attack against the British garrison at Bordentown, in order to block off any reinforcements. General James Ewing would take 700 militia across the river at Trenton Ferry, seize the bridge over the Assunpink Creek and prevent any enemy troops from escaping. | Where was there a British garrison located? | Bordentown | The American plan relied on launching coordinated attacks from three different directions. General John Cadwalder would launch a diversionary attack against the British garrison at $!Bordentown!$, in order to block off any reinforcements. General James Ewing would take 700 militia across the river at Trenton Ferry, seize the bridge over the Assunpink Creek and prevent any enemy troops from escaping. |

3 (b) Example of incorrect annotation: (Note: You don't need to add more context to the span)

| Paragraph | Question | Multi-choice Answer | Answer Span |
|---|---|---|---|
| Twentieth century research has shown that there are two pools of genetic variation: hidden and expressed. Mutation adds new genetic variation, and selection removes it from the pool of expressed variation. Segregation and recombination shuffle variation back and forth between the two pools with each generation. | Which process is responsible for adding genetic variation? | Mutation | Twentieth century research has shown that there are two pools of genetic variation: hidden and expressed. $!Mutation adds new genetic variation!$, and selection removes it from the pool of expressed variation. Segregation and recombination shuffle variation back and forth between the two pools with each generation. |

3 (b) In some cases, the answer can't be found in a concise form in the paragraph, thus requiring a longer span. Find the answer span where one can find the answer to the question, including the additional information that it may entail. Example of correct annotation where longer span is necessary:

| Paragraph | Question | Multi-choice Answer | Answer Span |
|---|---|---|---|
| This is called a chemical's pH. You can make an indicator using red cabbage juice. The cabbage juice changes color depending on how acidic or basic (alkaline) the chemical is. The pH level is indicated by the amount of Hydrogen (the H in pH) ions in the tested chemical. | How is the pH level of a chemical measured? | The amount of Hydrogen ions in the chemical | This is called a chemical's pH. You can make an indicator using red cabbage juice. The cabbage juice changes color depending on how acidic or basic (alkaline) the chemical is. The pH level is indicated by $!the amount of Hydrogen (the H in pH) ions in the tested chemical!$. |

**4. Finding a relevant span is more important than finding an exact answer.**

4 (a) In some cases, there is no exact answer in the paragraph, and therefore reasoning or deduction would be required to give a fully comprehensive answer. Yet keep in mind that the task is finding the span which points to *evidence for the answer*. In these cases, simply identify the span of text where an answer can be found. In the example below, the referent "this" can later be resolved by the reader by reading the earlier text in the passage (movement of men and materials).

| Paragraph | Question | Multi-choice Answer | Answer Span |
|---|---|---|---|
| Using ships to transport goods is by far the most efficient way to move large amounts of people and goods across oceans. The job of navies has traditionally been to ensure that your country maintains the ability to move your people and goods, while at the same time, interfering with your enemy's ability to move his people and goods. One of the most noteworthy recent examples of this was the North Atlantic campaign of WWII. The Americans were trying to move men and materials across the Atlantic Ocean to help Britain. At the same time, the German navy, using mainly U-boats, was trying to stop this traffic. Had the Allies failed, Germany probably would have been able to conquer Britain as it had the rest of Europe. | What was the German navy trying to accomplish during WWII? | Preventing Britain from receiving people and goods | Using ships to transport goods is by far the most efficient way to move large amounts of people and goods across oceans. The job of navies has traditionally been to ensure that your country maintains the ability to move your people and goods, while at the same time, interfering with your enemy's ability to move his people and goods. One of the most noteworthy recent examples of this was the North Atlantic campaign of WWII. The Americans were trying to move men and materials across the Atlantic Ocean to help Britain. At the same time, the German navy, using mainly U-boats, was trying to $!stop this traffic!$. Had the Allies failed, Germany probably would have been able to conquer Britain as it had the rest of Europe. |

**5. Don't include non-relevant punctuation**

5 (a) Final punctuation is usually not considered relevant

| Paragraph | Question | Multi-choice Answer | Answer Span |
|---|---|---|---|
| Things did not go well for the Italians in North Africa almost from the start. Within a week of Italy's declaration of war on June 10, 1940, the British 11th Hussars had seized Fort Capuzzo in Libya. In an ambush east of Bardia, the British captured the Italian Tenth Army's Engineer-in-Chief, General Lastucci. On June 28, Marshal Italo Balbo, the Governor-General of Libya and apparent heir to Mussolini, was killed by friendly fire while landing in Tobruk. | Where was Italo Balbo killed? | Tobruk | Things did not go well for the Italians in North Africa almost from the start. Within a week of Italy's declaration of war on June 10, 1940, the British 11th Hussars had seized Fort Capuzzo in Libya. In an ambush east of Bardia, the British captured the Italian Tenth Army's Engineer-in-Chief, General Lastucci. On June 28, Marshal Italo Balbo, the Governor-General of Libya and apparent heir to Mussolini, was killed by friendly fire while landing in $!Tobruk!$. |

5 (b) Closing punctuation is more likely to be relevant. ( ) " " [ ]
The example below includes the closing bracket in the span.

| Paragraph | Question | Multi-choice Answer | Answer Span |
|---|---|---|---|
| MRI is based on a physics phenomenon called nuclear magnetic resonance (NMR), which was discovered in the 1930s by Felix Bloch (working at Stanford University) and Edward Purcell (from Harvard University). In this resonance, magnetic field and radio waves cause atoms to give off tiny radio signals. In the year 1970, Raymond Damadian, a medical doctor and research scientist, discovered the basis for using magnetic resonance imaging as a tool for medical diagnosis. Four years later a patent was granted, which was the world's first patent issued in the field of MRI. In 1977, Dr. Damadian completed the construction of the first "whole-body" MRI scanner, which he called the "Indomitable". | Who discovered nuclear magnetic resonance? | Bloch and Purcell | MRI is based on a physics phenomenon called nuclear magnetic resonance (NMR), which was discovered in the 1930s by $!Felix Bloch (working at Stanford University) and Edward Purcell (from Harvard University)!$. In this resonance, magnetic field and radio waves cause atoms to give off tiny radio signals. In the year 1970, Raymond Damadian, a medical doctor and research scientist, discovered the basis for using magnetic resonance imaging as a tool for medical diagnosis. Four years later a patent was granted, which was the world's first patent issued in the field of MRI. In 1977, Dr. Damadian completed the construction of the first "whole-body" MRI scanner, which he called the "Indomitable". |

**6. In order to maintain consistency across annotators, the span will include the articles "the", "a" and "an".**

6 (a) An example of the inclusion of the article "the" in a span:

| Paragraph | Question | Multi-choice Answer | Answer Span |
|---|---|---|---|
| Golf is a game in which players use clubs to hit balls into holes. Eighteen holes are played during a regular round, with players usually starting on the first hole on the course and finishing on the eighteenth. The player who takes the fewest strokes, or swings of the club, to complete the course wins. The game is played on grass, and the grass around the hole is mown shorter and called the green. | On a golf course, where is the grass cut shorter? | On the green | Golf is a game in which players use clubs to hit balls into holes. Eighteen holes are played during a regular round, with players usually starting on the first hole on the course and finishing on the eighteenth. The player who takes the fewest strokes, or swings of the club, to complete the course wins. The game is played on grass, and the grass around the hole is mown shorter and called $!the green!$. |

6 (b) Example of inclusion of the article "a" in a span:

| Paragraph | Question | Multi-choice Answer | Answer Span |
|---|---|---|---|
| The atom can be considered to be one of the fundamental building blocks of all matter. Its a very complex entity which consists, according to a simplified Bohr model, of a central nucleus orbited by electrons, somewhat similar to planets orbiting the sun - see Figure 1.1. The nucleus consists of two particles - neutrons and protons. Protons have a positive electric charge while neutrons have no charge. The electrons have a negative electric charge. | The particles that orbit the nucleus have which type of charge? | Negative charge | The atom can be considered to be one of the fundamental building blocks of all matter. Its a very complex entity which consists, according to a simplified Bohr model, of a central nucleus orbited by electrons, somewhat similar to planets orbiting the sun - see Figure 1.1. The nucleus consists of two particles - neutrons and protons. Protons have a positive electric charge while neutrons have no charge. The electrons have $!a negative electric charge!$. |

**7. If the question is a When/ Where/ How question – then include the preposition or adverb that would normally form part of an appropriate answer.**

7 (a) An example of how-question where the preposition should be included in the span.

| Paragraph | Question | Multi-choice Answer | Answer Span |
|---|---|---|---|
| Subcultures bring together like-minded individuals who feel neglected by societal standards and allow them to develop a sense of identity. Subcultures can be distinctive because of the age, ethnicity, class, location, and/or gender of the members. The qualities that determine a subculture as distinct may be linguistic, aesthetic, religious, political, sexual, geographical, or a combination of factors. Members of a subculture often signal their membership through a distinctive and symbolic use of style, which includes fashions, mannerisms, and argot. | How do members of a particular subculture often signify their association with the group? | By using style as a form of symbolism | Subcultures bring together like-minded individuals who feel neglected by societal standards and allow them to develop a sense of identity. Subcultures can be distinctive because of the age, ethnicity, class, location, and/or gender of the members. The qualities that determine a subculture as distinct may be linguistic, aesthetic, religious, political, sexual, geographical, or a combination of factors. Members of a subculture often signal their membership $!through a distinctive and symbolic use of style!$, which includes fashions, mannerisms, and argot. |

7 (b) An example of a when-question where the preposition should be included in the span.

| Paragraph | Question | Multi-choice Answer | Answer Span |
|---|---|---|---|
| Gothic style peaked in the period between the 10th - 11th centuries and the 14th century. At the beginning dress was heavily influenced by the Byzantine culture in the east. However, due to the slow communication channels, styles in the west could lag behind by 25 to 30 year. Towards the end of the Middle Ages western Europe began to develop their own style. one of the biggest developments of the time as a result of the crusades people began to use buttons to fasten clothing. | When did western Europe stop relying heavily on influences and start developing its own style? | Around the end of the Middle Ages | Gothic style peaked in the period between the 10th - 11th centuries and the 14th century. At the beginning dress was heavily influenced by the Byzantine culture in the east. However, due to the slow communication channels, styles in the west could lag behind by 25 to 30 year. $!towards the end of the Middle Ages!$ western Europe began to develop their own style. one of the biggest developments of the time as a result of the crusades people began to use buttons to fasten clothing. |

7 (c) An example of a how-question case where the preposition should be included in the span.

| Paragraph | Question | Multi-choice Answer | Answer Span |
|---|---|---|---|
| The invention of spoke wheels made Assyrian chariots lighter, faster, and better prepared to outrun soldiers and other chariots. Arrows from their deadly crossbows could penetrate the armor of rival soldiers. About 1000 B.C., the Assyrians introduced the first cavalry. A cavalry is an army that fights on horseback. The saddle had not yet been invented, so the Assyrian cavalry fought on the bare backs of their horses. | How are battles that utilize a calvary fought? | on horseback | The invention of spoke wheels made Assyrian chariots lighter, faster, and better prepared to outrun soldiers and other chariots. Arrows from their deadly crossbows could penetrate the armor of rival soldiers. About 1000 B.C., the Assyrians introduced the first cavalry. A cavalry is an army that fights $!on horseback!$. The saddle had not yet been invented, so the Assyrian cavalry fought on the bare backs of their horses. |

**8. In some cases, the exact original answer string can be found multiple times in the paragraph. Choose the appropriate one only.**

| Paragraph | Question | Multi-choice Answer | Answer Span |
|---|---|---|---|
| Eighteen percent of Venezuelans are unemployed, and most of those who are employed work in the informal economy. Two thirds of Venezuelans who work do so in the service sector, nearly a quarter work in industry and a fifth work in agriculture. An important industry for Venezuelans is oil, where the country is a net exporter, even though only one percent work in the oil industry. | Which line of work mentioned in the passage employs the least number of Venezuelans? | Oil | Eighteen percent of Venezuelans are unemployed, and most of those who are employed work in the informal economy. Two thirds of Venezuelans who work do so in the service sector, nearly a quarter work in industry and a fifth work in agriculture. An important industry for Venezuelans is oil, where the country is a net exporter, even though only one percent work in the $!oil!$ industry. |

**9. The Belebele dataset was initially curated as a multiple-choice reading comprehension dataset. As a result, the questions were not crafted in a way that is suitable for extractive QA, and as such it may not be possible to find an exact answer or *any* answer:**

9 (a) You may come across some answers that cannot be found simply in a span of text i.e. reasoning or deduction is needed through reading the rest of the passage. In these cases, mark the span where evidence for the answer can be found. Then mark it as BB to indicate that it's problematic due to the nature of the Belebele dataset.

In some cases, you may need to unhide the original answer (column H) to help you decide whether or not it should be labelled it as BB.

Example of BB case:

| Paragraph | Question | Original Answer | Multi-choice Answer | Belebele Problem? |
|---|---|---|---|---|
| The balance of power was a system in which European nations sought to maintain the national sovereignty of all European states. The concept was that all European nations had to seek to prevent one nation from becoming powerful, and thus national governments often changed their alliances in order to maintain the balance. The War of Spanish Succession marked the first war whose central issue was the balance of power. This marked an important change, as European powers would no longer have the pretext of being religious wars. Thus, the Thirty Years' War would be the last war to be labeled a religious war. | In which country was the first war in Europe whose central issue was said to relate to the balance of power rather than having a religious context? | Spain | The balance of power was a system in which European nations sought to maintain the national sovereignty of all European states. The concept was that all European nations had to seek to prevent one nation from becoming powerful, and thus national governments often changed their alliances in order to maintain the balance. The War of Spanish Succession marked the first war whose central issue was the balance of power. This marked an important change, as European powers would no longer have the pretext of being religious wars. Thus, the Thirty Years' War would be the last war to be labeled a religious war. | BB |

9 (b) You may also be asked a true or false question. These are also not useful for our QA dataset. Remove the passage text and mark with X.

| Paragraph | Question | Multi-choice Answer | Answer Span |
|---|---|---|---|
| Virtual teams are held to the same standards of excellence as conventional teams, but there are subtle differences. Virtual team members often function as the point of contact for their immediate physical group. They often have more autonomy than conventional team members as their teams may meet according to varying time zones which may not be understood by their local management. The presence of a true "invisible team" (Larson and LaFasto, 1989, p109) is also a unique component of a virtual team. The "invisible team" is the management team to which each of the members report. The invisible team sets the standards for each member. | Based on the passage, which statement regarding physical and virtual teams is not true? | Conventional teams are usually held to a higher standard | X |

9 (c) In some cases, inference or calculation is required to establish the answer to the question. In other words, there is no particular span in which the answer can be found. Remove the passage text and mark as X.

| Paragraph | Question | Multi-choice Answer | Answer Span |
|---|---|---|---|
| MRI is based on a physics phenomenon called nuclear magnetic resonance (NMR), which was discovered in the 1930s by Felix Bloch (working at Stanford University) and Edward Purcell (from Harvard University). In this resonance, magnetic field and radio waves cause atoms to give off tiny radio signals. In the year 1970, Raymond Damadian, a medical doctor and research scientist, discovered the basis for using magnetic resonance imaging as a tool for medical diagnosis. Four years later a patent was granted, which was the world's first patent issued in the field of MRI. In 1977, Dr. Damadian completed the construction of the first "whole-body" MRI scanner, which he called the "Indomitable". | In what year was the first patent granted for medical imaging resonance? | 1974 | X |

# D Guidelines for Modern Standard Arabic Answer Span Annotation

## Belebele Answer Span: Arabic Annotation Guidelines

You'll be provided with a context paragraph and a question related to that paragraph. Your task is review and validate the pre-annotated *answer span* in the given paragraph. You should edit the span where necessary. The span indicates either the exact answer or where evidence of the answer can be found. The span should be as **concise as possible**.

The original Belebele multi-choice answer has been hidden in your file. However, the examples provided below include the multi-choice answer simply to illustrate the differences in the nature of the original answers and the "answer span" - which may help explain some annotation guideline decisions.

*********************************************************************************
**Please follow the annotation guidelines below carefully.**
*********************************************************************************

1. The span should identify the part of the text in which the answer to the question or evidence of the answer can be found
2. Where necessary, edit the start of the span with **$!** and the end of the span with **!$** . (Answer Span column V)
3. The answer span needs to be as short as possible.
4. Finding a relevant span is more important than finding an exact answer.
5. Don't include non-relevant punctuation.
6. Include the preposition if the question is a When/ How question.
7. In cases of multiple instances of the answer string, choose the appropriate one only.
8. Mark problematic questions with BB or X.

### Examples expanding on guidelines 3-8 are presented next.

**3. The answer span needs to be as short as possible.**

3 (a) Example of correct annotation:

| Paragraph | Question | Multi-choice Answer | Answer Span |
|---|---|---|---|
| تنتج النباتات الأكسجين الذي يتنفسه البشر، ويمتصون ثاني أكسيد الكربون الذي يخلفه البشر (أي الزفير). تصنع النباتات طعامها من الشمس من خلال التمثيل الضوئي. كما يوفرون الظل. نصنع بيوتنا وملابسنا من النباتات . معظم الطعام الذي نتناوله من النباتات كذلك. بدون النباتات، لا يمكن للحيوانات البقاء. | ما الذي يساعد النباتات في عملية التمثيل الضوئي؟ | الشمس | تنتج النباتات الأكسجين الذي يتنفسه البشر، ويمتصون ثاني أكسيد الكربون الذي يخلفه البشر (أي الزفير). تصنع النباتات طعامها من $!الشمس!$ من خلال التمثيل الضوئي. كما يوفرون الظل. نصنع بيوتنا وملابسنا من النباتات . معظم الطعام الذي نتناوله من النباتات كذلك. بدون النباتات، لا يمكن للحيوانات البقاء. |

3 (b) Example of incorrect annotation: (Note: You don't need to add more context to the span)

| Paragraph | Question | Multi-choice Answer | Answer Span |
|---|---|---|---|
| العضو الرئيسي للجهاز الدوري هو القلب الذي يضخ الدم. يتحرك الدم بعيداً عن القلب في أنابيب تسمى الشرايين ويعود إلى القلب في أنابيب تسمى الأوردة. أصغر الأنابيب تسمى الشعيرات الدموية. | ما هي الأنابيب الأصغر في جهاز الدورة الدموية؟ | الشعيرات الدموية | العضو الرئيسي للجهاز الدوري هو القلب الذي يضخ الدم. يتحرك الدم بعيداً عن القلب في أنابيب تسمى الشرايين ويعود إلى القلب في أنابيب تسمى الأوردة. أصغر الأنابيب تسمى $!الشعيرات الدموية.!$ |

3 (c) In some cases, the answer can't be found in a concise form in the paragraph, thus requiring a longer span. Find the answer span where one can find the answer to the question, including the additional information that it may entail. Example of correct annotation where longer span is necessary:

| Paragraph | Question | Multi-choice Answer | Answer Span |
|---|---|---|---|
| ضرب زلزال متوسط القوة مونتانا الغربية الساعة ١٠:٠٨ مساءً. يوم الاثنين. ولم ترد تقارير فورية عن وقوع أضرار من قبل هيئة المسح الجيولوجي الأمريكي (USGS) ومركز معلومات الزلازل الوطني التابع لها. وكان مركز الزلزال على بعد حوالي 20 كم (15 ميلاً) بين شمال وشمال شرق ديلون، وحوالي 65 كم (40 ميلاً) جنوب بوتي. | أين تمركز الزلزال بالنسبة لبوتي؟ | 40ميلا جنوباً | ضرب زلزال متوسط القوة مونتانا الغربية الساعة ١٠:٠٨ مساءً. يوم الاثنين. ولم ترد تقارير فورية عن وقوع أضرار من قبل هيئة المسح الجيولوجي الأمريكي (USGS) ومركز معلومات الزلازل الوطني التابع لها. وكان مركز الزلزال على بعد حوالي 20 كم (15 ميلاً) بين شمال وشمال شرق ديلون، **$!وحوالي 65 كم (40 ميلاً) جنوب بوتي!$**. |

**4. Finding a relevant span is more important than finding an exact answer.**

4 (a) In some cases, there is no exact answer in the paragraph, and therefore reasoning or deduction would be required to give a fully comprehensive answer. Yet keep in mind that the task is finding the span which points to *evidence for the answer*. In these cases, simply identify the span of text where an answer can be found. In the example below, the referent "this" can later be resolved by the reader by reading the earlier text in the passage (movement of men and materials).

| Paragraph | Question | Multi-choice Answer | Answer Span |
|---|---|---|---|
| قد يكون للطفرات مجموعة متنوعة من الآثار المختلفة، على حسب نوع الطفرة، وأهميّة قطعة المادّة الوراثيّة المتأثّرة، وما إذا كانت الخلايا المتأثرة من خلايا الخط الإنتاشي. يمكن فقط نقل الطفرات في خلايا الخط الجرثومي إلى الأطفال، بينما يمكن أن تسبب الطفرات في أماكن أخرى موت الخلايا أو السرطان. | على ماذا تعتمد قدرة الطفرة في الانتقال إلى الذرية؟ | إذا كانت الخلايا من خلايا الخط الإنتاشي | قد يكون للطفرات مجموعة متنوعة من الآثار المختلفة، على حسب نوع الطفرة، وأهميّة قطعة المادّة الوراثيّة المتأثّرة، وما إذا كانت الخلايا المتأثرة من خلايا الخط الإنتاشي. **$!يمكن فقط نقل الطفرات في خلايا الخط الجرثومي إلى الأطفال!$**، بينما يمكن أن تسبب الطفرات في أماكن أخرى موت الخلايا أو السرطان. |

**5. Don't include non-relevant punctuation**

5 (a) Final punctuation is usually not considered relevant

| Paragraph | Question | Multi-choice Answer | Answer Span |
|---|---|---|---|
| العضو الرئيسي للجهاز الدوري هو القلب الذي يضخ الدم. يتحرك الدم بعيداً عن القلب في أنابيب تسمى الشرايين ويعود إلى القلب في أنابيب تسمى الأوردة. أصغر الأنابيب تسمى الشعيرات الدموية. | أي جزء من الدورة الدموية يجلب الدم نحو القلب؟ | الأوردة | العضو الرئيسي للجهاز الدوري هو القلب الذي يضخ الدم. يتحرك الدم بعيداً عن القلب في أنابيب تسمى الشرايين ويعود إلى القلب في أنابيب تسمى **$!الأوردة!$**. أصغر الأنابيب تسمى الشعيرات الدموية. |

5 (b) Closing punctuation is more likely to be relevant. ( ) " " [ ]
The example below includes the closing bracket in the span.

| Paragraph | Question | Multi-choice Answer | Answer Span |
|---|---|---|---|
| "يعتمد التصوير بالرنين المغناطيسي على ظاهرة فيزيائية تسمى الرنين المغناطيسي النووي (NMR)، والتي اكتشفت في ثلاثينيات القرن الماضي عن طريق فيليكس بلوخ (من جامعة ستانفورد) وإدوارد بورسيل (من جامعة هارفارد) في هذا الرنين، تجعل المجالات المغناطيسيّة والموجات الراديويّة الذرّات تبعث إشارات راديو دقيقة. في عام 1970، اكتشف رايموند داماديان، وهو طبيب وعالم أبحاث، الأساس الذي بُني عليه استخدام التصوير بالرنين المغناطيسي كأداة للتشخيص الطبّي. بعد أربع سنوات، تم منح براءة اختراع، والتي كانت أول براءة اختراع يتم منحها في العالم في مجال التصوير بالرنين المغناطيسي. في عام 1977 استكمل الدكتور داماديان إنشاء أول ماسح ضوئي للتصوير بالرنين المغناطيسي ""لكامل الجسم""، والذي يسمى اليوم بـ""الذي لا يقهر"".". | من اكتشف الرنين المغناطيسي النووي؟ | بلوخ وبورسيل | "يعتمد التصوير بالرنين المغناطيسي على ظاهرة فيزيائية تسمى الرنين المغناطيسي النووي (NMR)، والتي اكتشفت في ثلاثينيات القرن الماضي عن طريق $!فيليكس بلوخ (من جامعة ستانفورد) وإدوارد بورسيل (من جامعة هارفارد)!$ في هذا الرنين، تجعل المجالات المغناطيسيّة والموجات الراديويّة الذرّات تبعث إشارات راديو دقيقة. في عام 1970، اكتشف رايموند داماديان، وهو طبيب وعالم أبحاث، الأساس الذي بُني عليه استخدام التصوير بالرنين المغناطيسي كأداة للتشخيص الطبّي. بعد أربع سنوات، تم منح براءة اختراع، والتي كانت أول براءة اختراع يتم منحها في العالم في مجال التصوير بالرنين المغناطيسي. في عام 1977 استكمل الدكتور داماديان إنشاء أول ماسح ضوئي للتصوير بالرنين المغناطيسي ""لكامل الجسم""، والذي يسمى اليوم بـ""الذي لا يقهر"".". |

**6. If the question is a When/ How question – then include the preposition or adverb that would normally form part of an appropriate answer. Note that sometimes the preposition is already part of the word in the form of a morpheme.**

6 (a) An example of a "how-question" where the preposition should be included in the span.

| Paragraph | Question | Multi-choice Answer | Answer Span |
|---|---|---|---|
| استطاع اختراع العجلات ذات العوارض أن يجعل العربات الأشورية أخف وزناً وأسرع وأكثر استعداداً لتجاوز الجنود والعربات الأخرى. يمكن أن تخترق السهام المُطلقة من الأقواس القاتلة دروع الجنود المتنافسين. قدم الأشوريون أول سلاح فرسان في حوالي عام 1000 قبل الميلاد. سلاح الفرسان هو جيش يقاتل على ظهر الخيل. لم يكن السرج قد اخترع بعد، لذلك قاتل الفرسان الأشوريون على ظهور خيولهم العارية. | **كيف** كانت تُخاض المعارك التي تستخدم الفرسان؟ | على ظهر الخيل | استطاع اختراع العجلات ذات العوارض أن يجعل العربات الأشورية أخف وزناً وأسرع وأكثر استعداداً لتجاوز الجنود والعربات الأخرى. يمكن أن تخترق السهام المُطلقة من الأقواس القاتلة دروع الجنود المتنافسين. قدم الأشوريون أول سلاح فرسان في حوالي عام 1000 قبل الميلاد. سلاح الفرسان هو جيش يقاتل $!على ظهور الخيل!$. لم يكن السرج قد اخترع بعد، لذلك قاتل الفرسان الأشوريون على ظهور خيولهم العارية.. |

6 (b) An example of a when-question where the preposition should be included in the span.

| Paragraph | Question | Multi-choice Answer | Answer Span |
|---|---|---|---|
| بلغ الطراز القوطي ذروته في فترة ما بين القرنين العاشر أو الحادي عشر إلى القرن الرابع عشر. في البداية، أثرت الثقافة البيزنطية في الشرق على الملابس تأثيرًا شديدًا. وعلى الرغم من ذلك، فنظرا لبطء قنوات التواصل، قد تتأخر الأنماط في الغرب قدر ما يقارب 25 إلى 30 سنة. في نهاية العصور الوسطى، بدأت أوروبا الغربية في تطوير أسلوبها الخاص. واحدة من أكبر التطورات في ذلك الوقت نتيجة للحروب الصليبية بدأ الناس في استخدام الأزرار لربط الملابس. | **متى** توقفت أوروبا الغربية عن الاعتماد الشديد على التأثيرات وبدأت في تطوير أسلوبها الخاص؟ | في نهاية العصور الوسطى | بلغ الطراز القوطي ذروته في فترة ما بين القرنين العاشر أو الحادي عشر إلى القرن الرابع عشر. في البداية، أثرت الثقافة البيزنطية في الشرق على الملابس تأثيرًا شديدًا. وعلى الرغم من ذلك، فنظرا لبطء قنوات التواصل، قد تتأخر الأنماط في الغرب قدر ما يقارب 25 إلى 30 سنة. $!في نهاية العصور الوسطى!$، بدأت أوروبا الغربية في تطوير أسلوبها الخاص. واحدة من أكبر التطورات في ذلك الوقت نتيجة للحروب الصليبية بدأ الناس في استخدام الأزرار لربط الملابس. |

**7. In some cases, the exact original answer string can be found multiple times in the paragraph. Choose the appropriate one only.**

| Paragraph | Question | Multi-choice Answer | Answer Span |
|---|---|---|---|
| "أفادت تقارير باستمرار أعمال النهب على نطاق واسع حتى صباح اليوم التالي وذلك لغياب ضباط إنفاذ القانون عن شوارع مدينة ""بشكيك"". وصف أحد المراقبين مدينة بشكيك بأنها تغرق في حالة من ""فقدان السيطرة""، حيث تجولت عصابات من الناس في الطرقات وقاموا بنهب مخازن السلع الاستهلاكية. حمّل العديد من سكان مدينة بشكيك المتظاهرين في الجنوب حالة الفوضى." | من ألقى باللوم على المتظاهرين من الجنوب في أعمال النهب؟ | سكان بيشكك | "أفادت تقارير باستمرار أعمال النهب على نطاق واسع حتى صباح اليوم التالي وذلك لغياب ضباط إنفاذ القانون عن شوارع مدينة ""بشكيك"". وصف أحد المراقبين مدينة بشكيك بأنها تغرق في حالة من ""فقدان السيطرة""، حيث تجولت عصابات من الناس في الطرقات وقاموا بنهب مخازن السلع الاستهلاكية. حمّل العديد من $!سكان مدينة ""بشكيك""!$ المتظاهرين في الجنوب حالة الفوضى." |

**8. The Belebele dataset was initially curated as a multiple-choice reading comprehension dataset. As a result, the questions were not crafted in a way that is suitable for QA, and as such it may not be possible to find an exact answer or *any* answer:**

8 (a) You may come across some answers that cannot be found simply in a span of text i.e. reasoning or deduction is needed through reading the rest of the passage. In these cases, mark the span where evidence for the answer can be found. Then mark it as BB to indicate that it's problematic due to the nature of the Belebele dataset.

Example of BB case:

| Paragraph | Question | Original Answer | Answer Span | Belebele Problem? |
|---|---|---|---|---|
| كانت الموازنة في الطاقة نظاماً حاولت من خلاله الدول الأوروبية أن تُحافظ على السيادة الوطنية الخاصة بجميع الدول الأوروبية. نص المفهوم على أن جميع الدول الأوروبية يجب أن تسعى لمنع دولة واحدة من أن تصبح قوية، وبالتالي عادةً ما غيرت الحكومات الوطنية تحالفاتها للحفاظ على التوازن. كانت حرب الخلافة الإسبانية هي الحرب الأولى التي كانت قضيتها المركزية هي ميزان القوى. كان ذلك بمثابة تغييرٍ مهم، حيث لم يعد لدى القوى الأوروبية ذريعة لكونها حروباً دينية. وهكذا، ستكون حرب الثلاثين عاماً آخر حربٍ توصف بأنها حرب دينية. | في أي بلد حصلت الحرب الأولى في أوروبا والتي كانت قضيتها المركزية مرتبطة بتوازن القوى بدلاً من أن تكون في سياق ديني؟ | إسبانيا | كانت الموازنة في الطاقة نظاماً حاولت من خلاله الدول الأوروبية أن تُحافظ على السيادة الوطنية الخاصة بجميع الدول الأوروبية. نص المفهوم على أن جميع الدول الأوروبية يجب أن تسعى لمنع دولة واحدة من أن تصبح قوية، وبالتالي عادةً ما غيرت الحكومات الوطنية تحالفاتها للحفاظ على التوازن. كانت $!حرب الخلافة الإسبانية!$ هي الحرب الأولى التي كانت قضيتها المركزية هي ميزان القوى. كان ذلك بمثابة تغييرٍ مهم، حيث لم يعد لدى القوى الأوروبية ذريعة لكونها حروباً دينية. وهكذا، ستكون حرب الثلاثين عاماً آخر حربٍ توصف بأنها حرب دينية | BB |

8 (b) Questions for which an answer (or evidence for an answer) could not be found at all have been marked as X (as per the English annotations)

Example of X case:

| Paragraph | Question | Original Answer | Answer Span | Belebele Problem? |
|---|---|---|---|---|
| كانت الموازنة في الطاقة"يعد الهرم الأكبر في محافظة ""الجيزة"" هو الوحيد من العجائب السبع التي لا تزال قائمة حتى يومنا هذا. تم بناؤه من قبل المصريين في القرن الثالث من الحقبة الحالية، يعتبر الهرم الأعظم واحداً من الأهرام الضخمة المبنية على شرف الفرعون الميت. تحتوي هضبة الجيزة، أو ""مدينة الجيزة الجنائزية"" بوادي الموتى المصري، على عدة أهرامات (أكبرها الهرم الأكبر) ومقابر صغيرة ومعابد وأبو الهول. تم إنشاء الهرم الأكبر لتكريم الفرعون خوفو، وتم بناء العديد من الأهرامات والمقابر والمعابد الأصغر لتكريم زوجات خوفو وأفراد أسرته." | ماذا يمكن العثور عليه في هضبة الجيزة؟ | جميع العجائب السبع | كانت الموازنة في الطاقة"يعد الهرم الأكبر في محافظة ""الجيزة"" هو الوحيد من العجائب السبع التي لا تزال قائمة حتى يومنا هذا. تم بناؤه من قبل المصريين في القرن الثالث من الحقبة الحالية، يعتبر الهرم الأعظم واحداً من الأهرام الضخمة المبنية على شرف الفرعون الميت. تحتوي هضبة الجيزة، أو ""مدينة الجيزة الجنائزية"" بوادي الموتى المصري، على عدة أهرامات (أكبرها الهرم الأكبر) ومقابر صغيرة ومعابد وأبو الهول. تم إنشاء الهرم الأكبر لتكريم الفرعون خوفو، وتم بناء العديد من الأهرامات والمقابر والمعابد الأصغر لتكريم زوجات خوفو وأفراد أسرته." | X |



\end{center}

\end{document}